%% file: main.tex
\documentclass[10pt,twocolumn,letterpaper]{article}

\usepackage{wacv}              

\usepackage{graphicx}
\usepackage{booktabs}
\usepackage{mathrsfs}
\usepackage{soul}
\usepackage{graphicx}
\usepackage{amsmath}
\usepackage{amssymb}
\usepackage{booktabs}
\usepackage{pifont}
\usepackage{array}
\usepackage[table]{xcolor} 
\usepackage[accsupp]{axessibility}  

\usepackage[pagebackref,breaklinks,colorlinks]{hyperref}

\usepackage[capitalize]{cleveref}
\crefname{section}{Sec.}{Secs.}
\Crefname{section}{Section}{Sections}
\Crefname{table}{Table}{Tables}
\crefname{table}{Tab.}{Tabs.}

\newcommand\salman[1]{\textcolor{black}{#1}}

\newcommand{\round}[1]{\left\lfloor#1\right\rceil}

\newcommand\blfootnote[1]{%
  \begingroup
  \renewcommand\thefootnote{}\footnote{#1}%
  \addtocounter{footnote}{-1}%
  \endgroup
}

\def\wacvPaperID{794} 
\def\confName{WACV}
\def\confYear{2025}

\begin{document}

\title{ELMGS: Enhancing memory and computation scaLability through coMpression for 3D Gaussian Splatting}



\author{ \hspace{-3.5mm}
\textbf{Muhammad Salman Ali}$^{1,2}$, \textbf{Sung-Ho Bae}$^{*2}$, \textbf{Enzo Tartaglione}\thanks{Corresponding Authors}$^{*1}$\\ 
$^1$LTCI, T\'el\'ecom Paris, Institut Polytechnique de Paris, France \\
$^2$ Kyung Hee University, Republic of Korea\\
\texttt{\{salmanali, shbae\}@khu.ac.kr,}\\
\texttt{enzo.tartaglione@telecom-paris.fr}
}

\maketitle

\begin{abstract}
3D models have recently been popularized by the potentiality of end-to-end training offered first by Neural Radiance Fields and most recently by 3D Gaussian Splatting models. The latter has the big advantage of naturally providing fast training convergence and high editability. However, as the research around these is still in its infancy, there is still a gap in the literature regarding the model's scalability. In this work, we propose an approach enabling both memory and computation scalability of such models. More specifically, we propose an iterative pruning strategy that removes redundant information encoded in the model. We also enhance compressibility for the model by including in the optimization strategy a differentiable quantization and entropy coding estimator. Our results on popular benchmarks showcase the effectiveness of the proposed approach and open the road to the broad deployability of such a solution even on resource-constrained devices.\blfootnote{This article has been accepted for publication at the 2025 IEEE/CVF Winter Conference on Applications of Computer Vision (WACV 2025).}
  
\end{abstract}
\input{display_figures/teaser}

\input{1_introduction}
\input{2_sota}

\input{3_method}
\input{4_results}
\input{5_conclusion}
\newpage
%
%
\bibliographystyle{splncs04}
\bibliography{main}
\end{document}

%% file: display_figures/teaser.tex
\begin{figure*}[t]
    \centering
    \includegraphics[width=\linewidth]{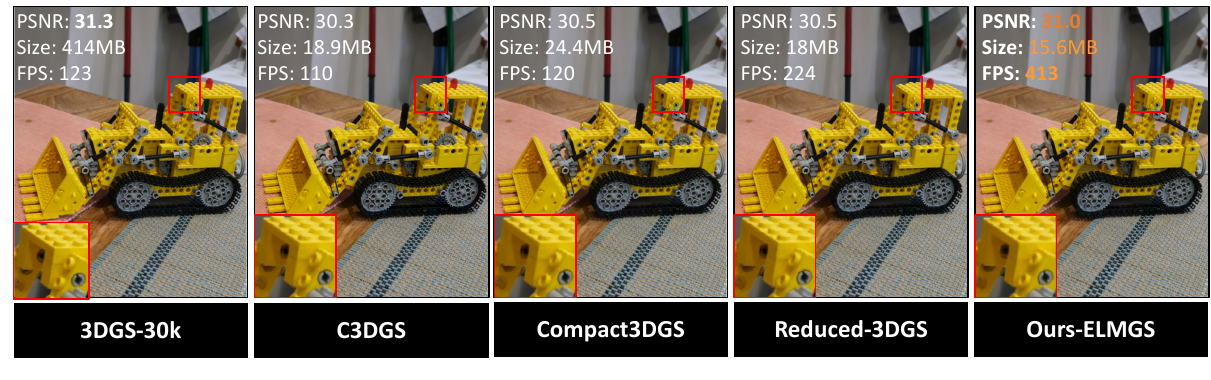}
    \vspace{-0.8cm}
    \caption{\salman{Qualitative comparison of ELMGS with 3DGS~\cite{kerbl20233d}, C3DGS~\cite{niedermayr2023compressed}, Compact3DGS~\cite{lee2023compact}, and Reduced-3DGS~\cite{papantonakis2024reducing}. With our proposed method we can achieve compression rates of about 27$\times$ with indiscernible loss in visual quality and significantly better rendering speed as compared to other methods. }}
    \label{fig:teaser}
    \vspace{-10pt}
\end{figure*}

%% file: 1_introduction.tex
\section{Introduction}
\label{sec:intro}
Recent advancements in novel view synthesis, driven by the emergence of Neural Radiance Fields (NeRFs)~\cite{mildenhall2021nerf}, have enabled the generation of new views of 3D scenes or objects by interpolating from a sparse collection of images with known camera angles. While NeRFs utilize an implicit neural representation to capture the volumetric radiance field for rendering novel views, they face significant limitations due to high memory requirements and computational complexity. This results in slow training and rendering times, prompting the exploration of alternative methods that often involve a trade-off between quality and complexity.

Using 3D voxel grids on the GPU combined with a multiresolution hash encoding of the input~\cite{muller2022instant} significantly reduces the operations needed and permits real-time performance. However, NeRFs still require high computational complexity due to which they cannot be deployed on edge devices with limited computational resources. Recent research has focused on reducing the memory footprint of NeRF by compressing the learned parametrizations on regular grids. These include vector quantized feature encoding~\cite{li2023compressing}, learned tensor decomposition~\cite{chen2022tensorf}, and frequency domain transformations~\cite{zhao2023tinynerf}.

Recently, a novel technique known as differentiable 3D Gaussian Splatting (3DGS) has been introduced to create a sparse adaptive representation of scenes, enabling high-speed rendering on GPUs. This method represents the scene as a collection of 3D Gaussians characterized by shape (covariance) and appearance (opacity, color, spherical harmonics) parameters, refined through differentiable rendering to align with a given set of images. This method offers a significant advantage over NeRF techniques in terms of both training and rendering speed. The efficiency stems from the straightforward process of projecting 3D Gaussians onto the 2D image space and subsequently rendering the view by combining multiple projected Gaussians with their opacity through rasterization. As a result, scenes can be rendered in real-time on a single GPU. Another notable benefit is that, unlike NeRF, the 3D scene's structure is explicitly stored in the parameter space rather than implicitly encoded in NeRF models.

Nevertheless, the optimized scenes generated by this approach frequently consist of millions of Gaussians, leading to parameters that are orders of magnitude larger than those in NeRFs. Consequently, the substantial storage and memory requirements make rendering a challenging task on devices with limited video memory, such as handheld devices or head-mounted displays. Although the specialized compute pipeline utilized for Gaussian rendering achieves real-time performance on high-end GPUs, the seamless integration of this pipeline into VR/AR environments or games, working in tandem with hardware rasterization of polygon models, presents challenges.

This paper focuses on enhancing the efficiency of Gaussian Splatting representations without compromising their fidelity, aiming to accelerate rendering speed for their utilization across diverse applications, including low-storage or low-memory IoT devices and AR/VR headsets. Our key insight is that within the final optimized scene, many Gaussians are redundant, characterized by opacity levels close to zero, rendering them ineffective. We begin with a pre-trained optimized Gaussian scene and prune it based on gradient and opacity levels, followed by fine-tuning to achieve superior performance compared to the baseline optimized scene. Subsequently, we employ quantization-aware training (QAT) to further compress the scene size. Upon completion of QAT, the scene undergoes entropy encoding, resulting in substantial compression gains along with performance improvements as shown in Fig.~\ref{fig:teaser}.

Our main contributions are listed here below.
\begin{itemize}
    \item We leverage the optimized Gaussians as a 3D prior for pruning, facilitating the removal of redundant Gaussians while fine-tuning the remaining ones to more accurately represent scene features~(\cref{sec:opaware}).
     \item Our compression pipeline achieves an improved balance between scene fidelity and compression surpassing the baseline~(\cref{sec:res}). 
    \item By integrating pruning with quantization-aware training, we enhance the compression of the scene representation and subsequently employ entropy coding for more efficient scene compression~(\cref{sec:qat} and~\cref{sec:ec}).
\end{itemize}

%% file: 2_sota.tex
\section{Related Works}
\label{sec:sota}

\input{display_figures/proposed_method}

\subsection{Novel View Synthesis}

Early methods for novel view synthesis using Convolutional Neural Networks (CNNs) encountered difficulties with Multi-View Stereo (MVS) geometry and temporal flickering \cite{flynn2016deepstereo,hedman2018deep}. The shift to volumetric representations began with Soft3D \cite{adams2017soft}, followed by techniques that integrated deep learning with volumetric raymarching~\cite{sitzmann2019deepvoxels}. NeRFs, proposed by Mildenhall \etal.~\cite{mildenhall2021nerf}, aimed to improve view synthesis quality but suffered from slow processing due to the use of a large Multi-Layer Perceptron (MLP) and dense sampling. Subsequent approaches like Mip-NeRF360~\cite{barron2022mip} sought a balance between quality and speed. Recent advancements focus on enhancing training and rendering speeds through optimized spatial data structures and MLP adjustments~\cite{chen2023mobilenerf, muller2022instant, hedman2021baking}, with methods like InstantNGP \cite{muller2022instant} and Plenoxels \cite{fridovich2022plenoxels} innovating for faster computations and eliminating neural networks, respectively. Despite progress, challenges persist in rendering speeds, image quality, and empty space representation. Conversely, 3DGS \cite{kerbl20233d} offers superior quality and speed without implicit learning~\cite{barron2022mip}.

\subsection{Differentiable Gaussian Splatting}
The Differentiable Gaussian splatting technique~\cite{kerbl20233d}, extends the EWA volume splatting method~\cite{zwicker2001ewa} to accurately compute the projections of 3D Gaussian kernels onto the 2D image plane. Furthermore, it employs differentiable rendering to iteratively optimize the number and parameters of the Gaussian kernels utilized for scene representation.

The target final scene representation consists of a collection of 3D Gaussians, each characterized by a covariance matrix \(\Sigma \in \mathbb{R}^{3\times 3}\) centered at location \(~{x \in \mathbb{R}^3}\). The covariance matrix can be parametrized by a rotation matrix \(R\) and a scaling matrix \(S\). For separate optimization of \(R\) and \(S\), Kerbl~\etal \cite{kerbl20233d} use a quaternion \(q\) to represent rotation and a vector \(s\) to represent scaling, both of which can be converted into their corresponding matrices. Additionally, each Gaussian possesses its opacity \(~{\alpha \in [0, 1]}\) and a set of spherical harmonics (SH) coefficients for reconstructing a view-dependent color.

To initialize the optimization process, 3DGS employs a point cloud obtained through a standard Structure from Motion (SfM) method~\cite{ullman1979interpretation}. During the training phase, 3DGS undertakes the rendering of training viewpoints and minimizes the loss between the ground truth and rendered images in the pixel space. The loss is a combination of $\ell_1$ loss and SSIM loss in the pixel space. Subsequently, it iteratively prunes points with small opacity parameters and introduces new ones when the gradient is considered ``substantial'' (where this is mediated through a thresholding function). The optimization involves adjusting the position ($x$), rotation ($q$), scaling ($s$), opacity ($\alpha$), and Spherical Harmonics (SH) coefficients of each 3D Gaussian to ensure that the rendered 2D Gaussians align with the training images. 3DGS showcases efficient training and real-time rendering performance, achieving or even exceeding the visualization quality benchmarks set by leading NeRF techniques~\cite{kerbl20233d}.

\subsection{Compression of 3D Gaussian Splatting}
\salman{The unstructured nature of 3DGS presents challenges for compression compared to NeRFs~\cite{chen2024hac, fei20243d}. Recent research has introduced structural modifications to improve compressibility~\cite{lu2023scaffold, chen2024hac}. For example, Scaffold-GS employs anchor points to distribute local 3D Gaussians and adjusts their attributes according to viewing direction and distance within the view frustum. The Hash-grid Assisted Context (HAC)~\cite{chen2024hac} framework builds on Scaffold-GS by co-learning a compact hash grid that models the context of anchor attributes. RadSplat~\cite{niemeyer2024radsplat} uses NeRFs as a prior for optimizing 3DGS and incorporates a pruning technique to achieve a compact scene representation. MiniSplatting~\cite{fang2024mini} addresses the inefficient spatial distribution of Gaussians by introducing densification strategies, such as blur split and depth reinitialization, leading to a more uniform and efficient spatial arrangement.}

Niedermayr\etal~\cite{niedermayr2023compressed} proposed a compression framework that maintains the original 3DGS parameters while compressing directional colors and Gaussian parameters. This framework uses sensitivity-aware vector clustering for pruning and QAT, achieving compression rates up to 30$\times$ with negligible performance loss relative to the original 3DGS. Likewise, Compact3DGS~\cite{lee2023compact} employs a learnable mask to prune Gaussians and uses a grid-based neural field to compactly represent view-dependent colors, replacing spherical harmonics. Additionally, Compact3DGS leverages vector quantization to efficiently encode the geometric attributes of Gaussians into codebooks.

Previous research on 3DGS compression has explored vector quantization schemes~\cite{niedermayr2023compressed,lee2023compact}, which, while beneficial for compression, pose challenges for hardware implementation. Previous work in compression has shown that uniform quantization schemes are the most hardware-friendly~\cite{liu2022nonuniform, gholami2022survey}.
To the best of our knowledge, we are among the first works~\cite{chen2024hac} to propose a QAT method based on a scalar uniform quantization scheme.

Our method aims to enhance or replace existing masking strategies for insignificant Gaussians (see Sec.~\ref{sec:res}), provide a more hardware-compatible quantization scheme, and surpass the performance of current techniques. However, this paper focuses exclusively on the vanilla 3DGS variant, given its broad applicability.

%% file: display_figures/proposed_method.tex
\begin{figure*}[t]
    \centering
    \includegraphics[width=0.9\linewidth]{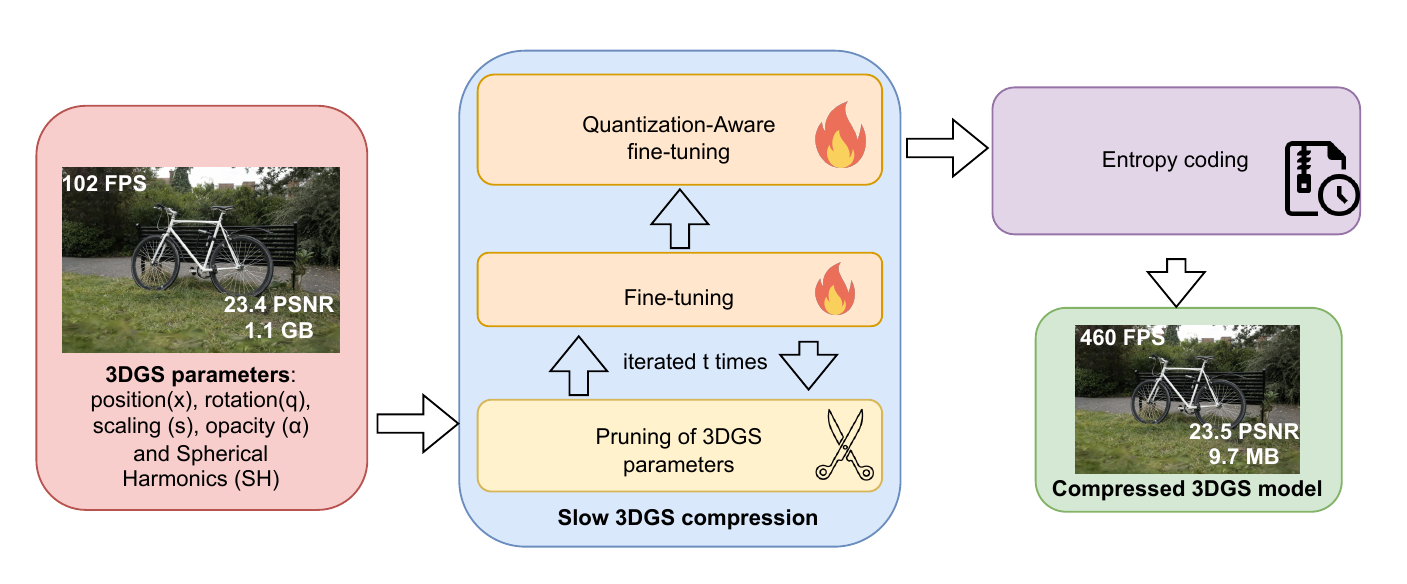}
    \caption{\salman{ELMGS begins with a pre-trained 3DGS scene and performs iterative pruning with finetuning to remove less significant Gaussians, followed by Quantization-Aware finetuning. The quantized model is then entropy-encoded to generate the final compressed scene.}}
    \label{fig:method}
\end{figure*}

%% file: 3_method.tex
\input{display_figures/iterative_pruning}
\input{display_figures/pruning_and_opacity}

\section{Methodology}
\label{sec:method}

3DGS models, despite showcasing remarkable performances and enhanced editability, unfortunately, fall short in storage memory, as they typically require massive memory. In our approach, we propose a simple yet effective strategy to compress such a model, that can be decomposed in three distinct steps (Fig.~\ref{fig:method}). The first one employs gradient and opacity-aware pruning (discussed in Sec.~\ref{sec:opaware}), where we iteratively remove parameters from the 3DGS model and fine-tune the model. Implicitly, this provides a strong prior to the geometry of the scene. Later, we perform the quantization-aware fine-tuning step (discussed in Sec.~\ref{sec:qat}), which prepares the field for the entropy coding step (detailed in Sec.~\ref{sec:ec}).

\subsection{Gradient and Opacity Aware Pruning (GAP)}
\label{sec:opaware}

3DGS typically requires several million Gaussians to effectively model a standard scene. Each Gaussian entails 59 parameters, resulting in a storage size considerably larger than most NeRF methodologies, such as Mip-NeRF360~\cite{barron2022mip}, K-planes~\cite{fridovich2023k} and instantNGP~\cite{muller2022instant}. This renders it inefficient for certain applications, particularly those involving edge devices. Our focus lies in parameter reduction. In the original training process of 3DGS, the authors pruned and densified Gaussians up to a specified number of iterations. Pruning was based on a predetermined opacity threshold. However, upon examination of the final optimized scene (see Fig.\ref{fig:abl-studies}c), a substantial number of Gaussians were found to fall below this opacity threshold, rendering them essentially redundant. \salman{Consequently, we adopt a pruning strategy based on the quantile functions of opacity and its corresponding gradient. The quantile function applies a threshold $(\gamma_{\text{iter}})$ to identify a percentage of parameters that fall below the threshold value $(\gamma_{\text{iter}})$. If both the opacity and gradient of a given Gaussian fall below the specified threshold,$(\gamma_{\text{iter}})$ the Gaussian can be pruned with minimal or no adverse impact on the rendered scene quality}. We can define GAP as: 
\begin{equation}
    \label{eq: pruning}
    \boldsymbol{\Sigma}' = \begin{cases}
    &~~~~\left[\left|\Sigma_i^\alpha\right|\geq \mathcal{Q}_{|\boldsymbol{\Sigma^\alpha}|}(\gamma_{\text{iter}})\right] \\
     \Sigma_i & \text{if }  ~~~~~~~~~~~~~~~~~\lor  \\ 
     &~~~~\left[ \left|\nabla\Sigma_i\right| \geq \mathcal{Q}_{|\boldsymbol{\nabla\Sigma}|}(\gamma_{\text{iter}}) \right] \\~\\
    0  & \text{otherwise,}
    \end{cases}
\end{equation}
\noindent where $\nabla\Sigma_i$ denotes the gradient for the $i$-th Gaussian, $\Sigma_i^\alpha$ denotes the $\alpha$ value of the $i$-th Gaussian, and $\mathcal{Q}_{|\boldsymbol{\Sigma^\alpha}|}(.)$ represents the quantile function for the opacity, $\mathcal{Q}_{|\boldsymbol{\nabla\Sigma}|}(.)$ is the quantile function for the gradients of the Gaussians, and $\gamma \in [0, 1]$ denotes the fraction of Gaussians to be removed.

This GAP pruning process, coupled with periodic finetuning, not only enhances performance but also yields significant compression gains. 
The pruning and fine-tuning strategy allows us to eliminate redundant Gaussians and refine the remaining ones to more accurately represent the scene compared to the baseline. We know from previous works on scene rendering that a low iterative pruning strategy enables access to much sparser models while still maintaining high fidelity~\cite{deng2023compressing}; however, it is still unclear what could it be the effect on 3DGS models. We argue that, given a target sparsity $\gamma_{\text{target}}$, employing a gradual pruning for $t$ iterations, and for instance applying a sparsification process on the 3DGS model applying 
\begin{equation}
    \gamma_{\text{iter}} = 1 - (1-\gamma_{\text{target}})^{\frac{1}{t}},
\end{equation}
would lead to enhanced results. This happens thanks to two effects. First, the fine-tuning process (that follows the pruning step) adjusts the covariance $\Sigma$ (still minimizing the rendering loss) such that solid surfaces will have higher values, while semi-transparent artifacts remain at lower values and can be removed at the next iteration. Second, the slow iterative process prevents the optimization algorithm from falling into a sub-optimal local minimum. Specifically for this reason, although a pruning mechanism is already employed in 3DGS, it is important to begin with an overparametrized but well-performing model. Some parallels can be drawn also from traditional deep learning~\cite{blalock2020state, woodworth2020kernel}. Fig.~\ref{fig:prune} pictures the effect of either performing a slow pruning or a one-shot one, where the target sparsity level is reached at once.

\subsection{Quantization-Aware Training (QAT)}
\label{sec:qat}
We also employ Learned Step Size Quantization (LSQ)~\cite{esser2020learned} to optimize the quantization mapping for all parameters in 3DGS. LSQ provides a direct way to approximate the gradient for the quantizer step size by accounting for quantized state transitions. This method facilitates more precise optimization by treating the step size as a model parameter. Furthermore, LSQ employs a straightforward heuristic to effectively balance step size updates with weight updates, making it well-suited for QAT on 3DGS.

Given parameter to quantize \(\Sigma\), quantizer step size \(\Delta\), the number of positive and negative quantization levels \(Q_P\) and \(Q_N\), respectively, we define a quantizer that computes \(\bar{\Sigma}\), a quantized and integer-scaled representation of the data, and \(\hat{\Sigma}\). This results in a quantized representation of the data at the same scale as \(\Sigma\):

\begin{equation}
\label{lsq eq:1}
    \begin{aligned}
        &\bar{\Sigma}=\left\lfloor \text{clip} \left(\frac{\Sigma}{\Delta},-Q_N, Q_P\right)\right\rceil,\\
    \end{aligned}
\end{equation}

\begin{equation}
\label{lsq eq:2}
    \begin{aligned}
        &\hat{\Sigma}=\bar{\Sigma} \cdot \Delta ,
    \end{aligned}
\end{equation}

where \(\text{clip}(z, r_1, r_2) \) adjusts \( z \) by setting values below \( r_1 \) to \( r_1 \) and values above \( r_2 \) to \( r_2 \). Additionally, \(\round{z} \) rounds \( z \) to the nearest integer. With \( b \)-bit encoding, for unsigned data, the feature values will range from \( Q_N = 0 \) to \( Q_P = 2^b - 1 \). For signed data, \( Q_N = -2^{(b-1)} \) and \( ~{Q_P = 2^{(b-1)} - 1} \).

 LSQ introduces a mechanism to learn the scale (\( \Delta \)) based on the training loss by incorporating the following gradient through the quantizer to the step size parameter:
\begin{equation}
\frac{\partial \hat{\Sigma}}{\partial \Delta}= 
    \begin{cases}
        -\frac{\Sigma}{\Delta}+\lfloor \frac{\Sigma}{\Delta}\rceil & \text { if }-Q_N<\frac{\Sigma}{\Delta}<Q_P \\ 
        -Q_N & \text { if } \frac{\Sigma}{\Delta} \leq-Q_N \\ 
        Q_P & \text { if } \frac{\Sigma}{\Delta} \geq Q_P.
    \end{cases}.
\end{equation}
This gradient is derived by employing the straight-through estimator~\cite{bengio2013estimating} to approximate the gradient through the round function as a pass-through operation (while retaining the round operation itself for the sake of differentiating downstream operations), and differentiating all other operations in \eqref{lsq eq:1} and \eqref{lsq eq:2} as usual.
\input{display_figures/ours_big_comp}
\subsection{Entropy Encoding}

\label{sec:ec}

GAP significantly reduces the number of Gaussians, which are then subjected to quantization-aware training (QAT) to quantize all Gaussian parameters. The resulting quantized feature representation of the Gaussians enables high compression rates using entropy coding (EC). Finally, the quantized Gaussians undergo compression using the LZ77~\cite{ziv1977universal} algorithm. By arranging the Gaussians according to their positions along a Z-order curve in Morton order (MO), we can further exploit coherence and enhance the effectiveness of entropy encoding. We can expect that our iterative pruning approach is also enhancing the model's compressibility, and for instance the entropy: this solves the pressing issue of including entropy encoding in the optimization loop (and bypassing practical obstacles including the differentiability of such an element). For instance, as suggested in Sec.~\ref{sec:opaware}, thanks to the pruning approach we expect high peaks around high-density values to rise naturally, showcasing a trend visualized in Fig.~\ref{fig:entropy}. Evidently, as the frequency around specific values increases, the first-order entropy of the encoded model (acting as an upper bound) is also minimized.

In the next section, we will show our empirical findings in the typical benchmarked setups.

%% file: display_figures/iterative_pruning.tex
\begin{figure*}[t]
    \centering
    \includegraphics[width=0.8\linewidth]{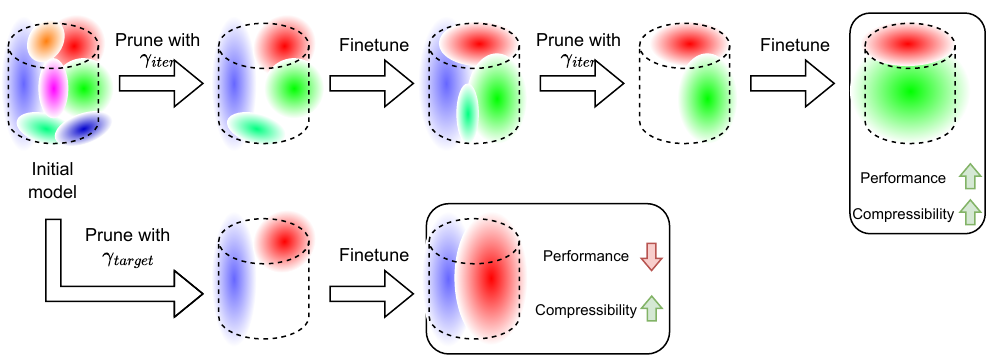}
    \caption{Effect of gradual pruning to the 3DGS model: a gradual removal allows the model to self-adjust and to better fit the scene/object.}
    \label{fig:prune}
\end{figure*}

%% file: display_figures/pruning_and_opacity.tex
\begin{figure}[t]
    \centering
    \includegraphics[width=\linewidth]{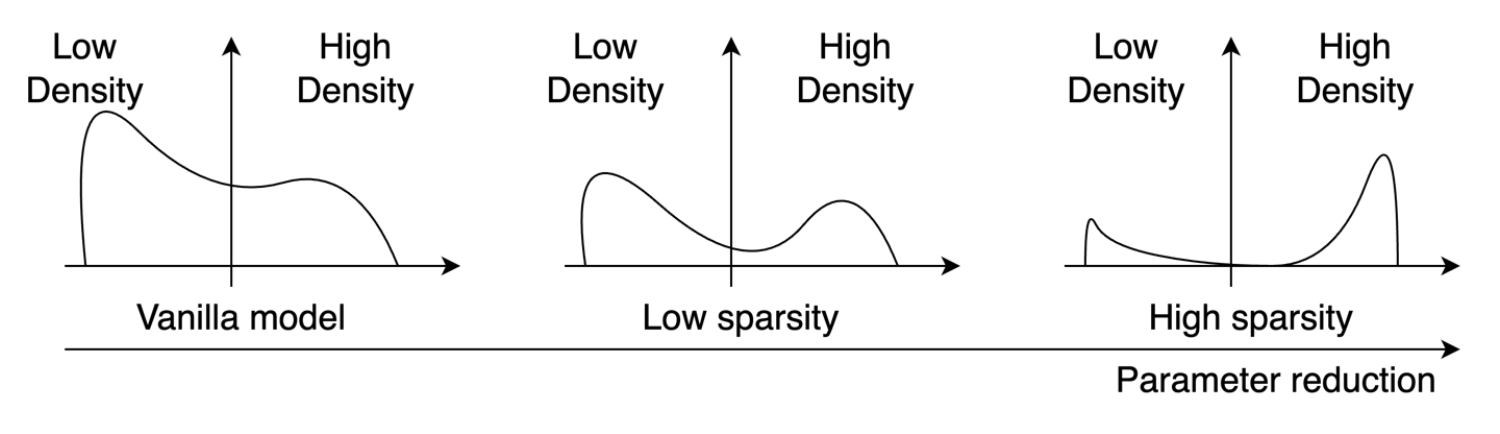}
    \caption{Effect of gradual pruning to the opacity values. We have two effects: (i) the number of parameters in our model reduces; (ii) the frequency around specific low and high density is increased.}
    \label{fig:entropy}
    \vspace{-0.4cm}
\end{figure}

%% file: display_figures/ours_big_comp.tex
\begin{figure*}[t]
\includegraphics[width=\linewidth]{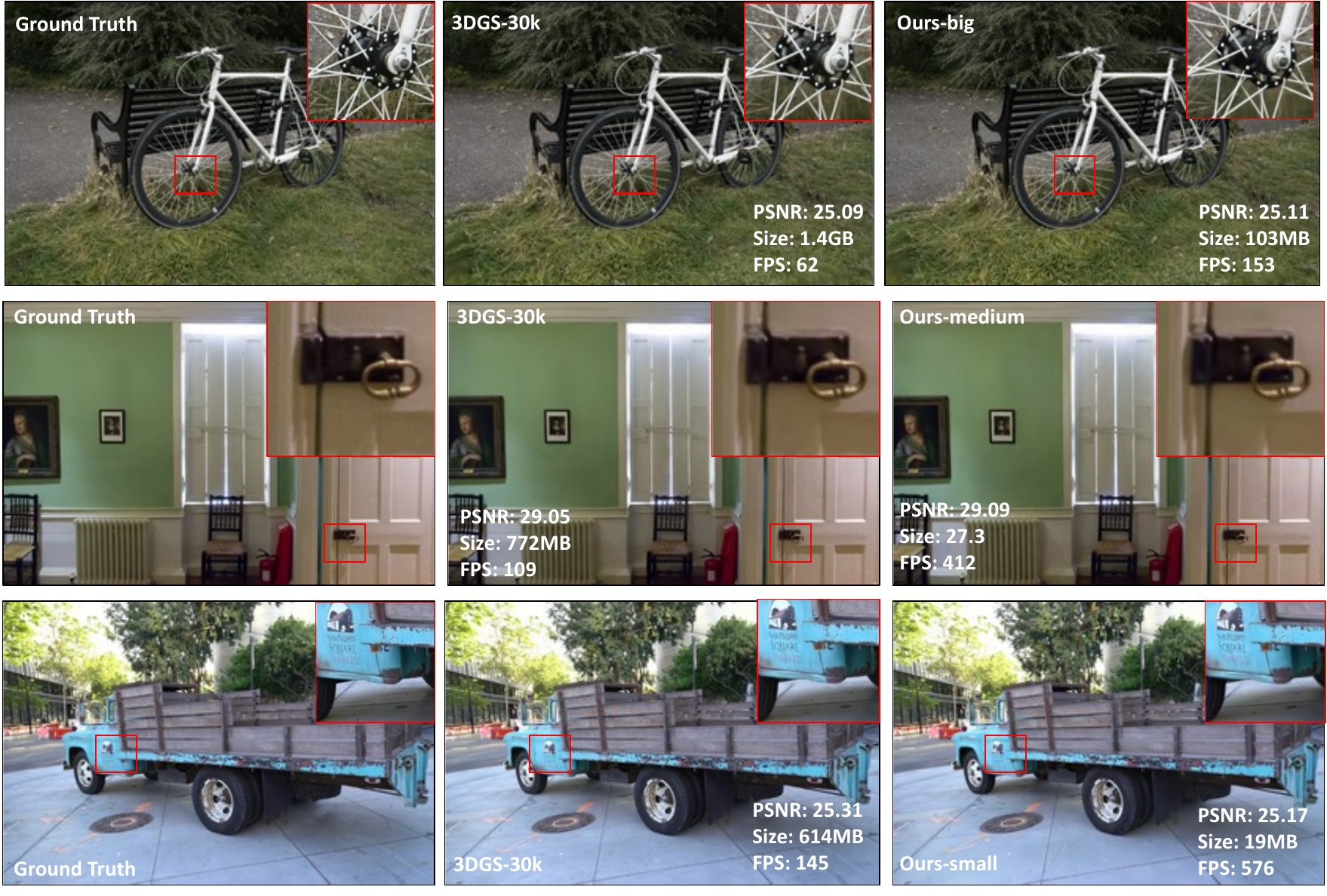}
    \centering
    \caption{Comparison of ground truth images from the test set of \texttt{bicycle},  \texttt{drjohnson}, and \texttt{truck} scenes between ELMGS ``ours-big", ``ours-medium" and ``ours-small" compressed representation and 3DGS-30k.}
    \label{fig: ours-big}
    \vspace{-0.2cm}
\end{figure*}

%% file: 4_results.tex
\section{Experiments and Results}
\label{sec:results}

\input{tables/comp_table}
\input{tables/pruning_level_comp}

In this section, we are going to present our empirical findings on typical well-established benchmarks in the 3DGS community. First, we will provide the implementation details for our employed method alongside an overview of the benchmarked datasets and the evaluation metrics (Sec.~\ref{sec:idf}); then we present the qualitative and quantitative results (Sec.~\ref{sec:res}); and finally, a detailed ablation study (Sec.~\ref{sec:abl}). More results are also presented in the Supplementary Material, alongside the code, which will be open-sourced upon acceptance of the article.

\subsection{Implementation Details}
\label{sec:idf}
For all our experiments, we utilize the publicly available official code repository of 3DGS~\cite{kerbl20233d} provided by its authors, without altering the hyperparameters used for training compared to 3DGS. We initiate pruning with the optimized Gaussians trained for 30,000 iterations and apply pruning using \eqref{eq: pruning} with $\gamma_{\text{iter}}$ where~$\gamma_{\text{iter}}\in[0.375,0.6]$. The model undergoes sequential pruning with the same $\gamma_{\text{iter}}$ value after every 500 iterations until 35,000 iterations, followed by further finetuning for 5,000 iterations. In most experiments, pruning and finetuning yield superior performance compared to the baseline. Subsequently, we train the model using QAT with the LSQ method and further finetune it for 5,000 iterations on QAT. For QAT, we encode Spherical Harmonics (SH) features into 8 bits, while all other features such as opacity, scaling, rotation, and XYZ positions are quantized into 32 bits. Finally, the finetuned model after QAT undergoes entropy coding to complete our compression pipeline. Our method extends the training time of 3DGS by half, totaling 45,000 iterations.

\noindent
\textbf{Datasets.} We evaluate the effectiveness of our compression and rendering method across various scenes, including the Mip-Nerf360 \cite{barron2022mip} indoor and outdoor scenes, as well as two scenes from the Tanks\&Temples \cite{knapitsch2017tanks} and Deep Blending \cite{hedman2018deep} datasets.

\noindent
\textbf{Evaluation.} To ensure a fair comparison, we maintain consistency with the train-test split used in Mip-NeRF360 \cite{barron2022mip} and 3DGS \cite{kerbl20233d}, and directly present the metrics for other methods as reported in \cite{kerbl20233d}. 
Our evaluation encompasses standard metrics such as SSIM, PSNR, and LPIPS, along with the FPS and average memory consumption across all datasets.

\subsection{Results}
\label{sec:res} \noindent
\textbf{Quantitative Comparison.} Table~\ref{tab:comp_sota} compares our method with existing state-of-the-art approaches. C3DGS~\cite{niedermayr2023compressed}, Compact3DGS~\cite{lee2023compact}, and Reduced-3DGS~\cite{papantonakis2024reducing} exhibit performance similar to the 3DGS-30k baseline, with a minor improvement in rendering speed. Only Reduced-3DGS~\cite{papantonakis2024reducing} shows a notable FPS increase, improving by 2.5$\times$. In contrast, our ``Ours-big" variant surpasses the 3DGS baseline, achieving a 14$\times$ compression ratio and a 2.2$\times$ increase in FPS. The ``Ours-medium" variant maintains baseline performance while offering a 22$\times$ compression ratio and a 3$\times$ increase in FPS. Notably, the ``Ours-small" variant achieves a remarkable 38$\times$ compression ratio and a 4$\times$ increase in rendering speed, averaging around 520 FPS across all datasets. Our method significantly reduces the model's memory footprint, making it competitive with NeRF approaches and addressing a key limitation of 3DGS models. The substantial compression is especially noteworthy given that many Gaussians in the original 3DGS are non-essential.

\noindent
\textbf{Qualitative Comparison.}
Fig. \ref{fig: ours-big} compares the ELMGS ``Ours-big", ``Ours-medium", and ``Ours-small" models with the 3DGS-30k baseline. For the \texttt{bicycle} scene, ``Ours-big" not only delivered better visual quality but also achieved a compression ratio of 14$\times$ with about 2.5$\times$ improvement in rendering speed. Similarly, ``Ours-medium" provided a better PSNR with a compression ratio of approximately 28$\times$ and an increase in rendering speed by about 4$\times$ on the \texttt{drjohnson} scene. ``Ours-small" achieved similar visual quality for the \texttt{truck} scene while achieving a compression ratio of about 32$\times$ and an FPS gain of about 4$\times$.
Please refer to the Supplementary Material for additional images and further qualitative comparisons.

\input{tables/modules_impact}
\input{display_figures/abalation_figures}
\subsection{Ablation Studies}
\label{sec:abl}
\noindent
\textbf{Effect of different Pruning levels.}
Table~\ref{tab:pruning_comp} illustrates the impact of different levels of pruning on the benchmark datasets. Across all datasets, it is evident that even with high pruning levels, our proposed method can still deliver reasonable performance. For the smallest Gaussian splats, the average PSNR is 25.7 with an average size of 6.4 MBs, while for the least compressed Gaussian splats, the average PSNR is 27.04 with an average size of 59 MBs. By varying the pruning levels, we are able to achieve significantly smaller Gaussian splats with good performance, paving the way for them to be deployed in smaller end-to-end devices where memory footprint is crucial.

\noindent
\textbf{Quantization Bits.}
Gaussian splats exhibit sensitivity to quantization bits, with only spherical harmonics (SH) features akin to those in deep neural networks that can be quantized to lower bit depths without loss in information. However, other attributes like opacity, scaling, rotation, and XYZ positions pose challenges for quantization due to their intricate nature. To explore the impact of varying quantization bit levels on the Tanks\&Temples dataset, we conducted experiments across different levels of quantization bits for various Gaussian parameters. The results are presented in Fig.~\ref{fig:abl-studies}a. For all the experiments, the spherical harmonics (SH) features were fixed at 8 bits. Fig.~\ref{fig:abl-studies}a illustrates a notable decline in performance as quantization bits decrease, while the compression ratio remains the same. Therefore, to minimize quantization errors, we adopted 32 bits as the standard setting for all experiments.

\noindent
\textbf{FPS Gain with ELMGS.}
ELMGS significantly boosts the FPS rate of 3DGS. On the Tanks\&Temples dataset, our method achieves an FPS of over 600 while maintaining state-of-the-art performance, as shown in Fig.~\ref{fig:abl-studies}b. The renderings were performed using an NVIDIA RTX-3090 using Kerb \etal~\cite{kerbl20233d} rasterization approach, with the final FPS averaged over three separate runs. Compared to Compact3DGS~\cite{lee2023compact}, C3DGS~\cite{niedermayr2023compressed}, and the 3DGS-30k baseline, we observe an approximate 4$\times$ and about 2$\times$ improvement compared to Reduced-3DGS~\cite{papantonakis2024reducing} in FPS while maintaining better/ similar performance.

\noindent
\textbf{Impact of each module on compression.}
Table~\ref{table:post-hoc-pruning} shows the performance and compression impact of each module within our proposed method. With the integration of GAP, our approach achieves a compression gain of 5$\times$, while the introduction of LSQ does not directly affect compression. However, employing Morton ordering (MO) and entropy encoding (EC) on quantized Gaussians yields substantial gains (22$\times$) compared to solely applying EC on pruned Gaussians (5.5$\times$) signifying the importance of each module in our proposed method. 

\noindent
\textbf{Impact of GAP on Opacity.}
GAP significantly reduces the number of Gaussians and enforces a strong prior on the scene. As depicted in Fig.~\ref{fig:abl-studies}c, the opacity distribution for the \texttt{garden} scene changes significantly before and after pruning. In the baseline 3DGS, most opacity values are very low, indicating minimal contribution to scene reconstruction. However, after GAP, a substantial proportion of opacity values increase significantly, indicating that the model learns a more solid geometry.

%% file: tables/comp_table.tex
\begin{table*}[t]
    \centering
    \caption{Comparison with SOTA methods for novel view synthesis. $^\dag$ Reported from \cite{kerbl20233d}. Red indicating the best performance, followed by yellow and green.}
    \label{tab:comp_sota}
    \scalebox{0.68}{
    \begin{tabular}{l|ccccc|ccccc|ccccc}
    \toprule
    & \multicolumn{5}{c}{Mip-NeRF360} & \multicolumn{5}{c}{Tanks\&Temples} & \multicolumn{5}{c}{Deep Blending} \\
    \textbf{Method} & \textbf{SSIM$^\uparrow$} & \textbf{LPIPS$^\downarrow$} & \textbf{PSNR$^\uparrow$} & \textbf{FPS$^\uparrow$} & \textbf{Mem$^\downarrow$} & \textbf{SSIM$^\uparrow$} & \textbf{LPIPS$^\downarrow$} & \textbf{PSNR$^\uparrow$} & \textbf{FPS$^\uparrow$} & \textbf{Mem$^\downarrow$} & \textbf{SSIM$^\uparrow$} & \textbf{LPIPS$^\downarrow$} & \textbf{PSNR$^\uparrow$} & \textbf{FPS$^\uparrow$} & \textbf{Mem$^\downarrow$} \\
    \midrule
    3DGS-7k$^\dag$     & 0.770 & 0.279 & 25.60 & 160 & 523MB &0.767 & 0.280 & 21.20 & 197 & 270MB & 0.875 & 0.317 & 27.78 & 172 & 386MB \\
    3DGS-30k$^\dag$     & 0.815 & 0.214 & \cellcolor{green!25}27.21 & 134 & 734MB & 0.841 & 0.183 & 23.14 & 154 & 411MB & 0.903 & 0.243 & 29.41 & 137 & 676MB \\
    \midrule
    C3DGS~\cite{niedermayr2023compressed}  & 0.801 & 0.238 & 26.98 & 113 & \cellcolor{yellow!25}28.8MB & 0.832 & 0.194 & 23.32 & 149 & \cellcolor{green!25} 17.3MB & 0.898 & 0.253 & 29.38 & 128 & 25.3MB \\
    Compact3DGS~\cite{lee2023compact} & 0.798 & 0.247 & 27.08 & 128 & 48.8MB & 0.831 & 0.201 & 23.32 & 185 & 39.4MB & 0.901 & 0.258 & \cellcolor{red!25}29.79 & 181 & 43.2MB \\
    Compact3DGS+PP~\cite{lee2023compact} & 0.797 & 0.247 & 27.03 & - & 29.1MB & 0.831 & 0.202 & 23.32 & - & 20.9MB & 0.900 & 0.258 & 29.73 & - & 23.8MB \\
    Reduced-3DGS~\cite{papantonakis2024reducing} & 0.809 & 0.226 & 27.10 & \cellcolor{green!25}284 & \cellcolor{green!25}29.0MB & 0.840 & 0.188 & 23.57 & \cellcolor{green!25} 433 & \cellcolor{yellow!25} 14.0MB & 0.902 & 0.249 & \cellcolor{yellow!25}29.63 & \cellcolor{green!25} 360 & \cellcolor{yellow!25}18.0MB \\
    \midrule

    \textbf{Ours-big} & 0.808 & 0.235 & \cellcolor{red!25}27.58 & 216 & 62.2MB & 0.845 & 0.191 & \cellcolor{yellow!25}24.06 & 418 & 27.9MB & 0.899 & 0.255 & \cellcolor{green!25}29.57 & 321 & 40.8MB \\
    \textbf{Ours-medium} & 0.792 & 0.264 & \cellcolor{yellow!25}27.31 & \cellcolor{yellow!25}290 & 38.6MB & 0.838 & 0.191 & \cellcolor{red!25}24.08 & \cellcolor{yellow!25}600 & 18.8MB & 0.897 & 0.261 & 29.48 & \cellcolor{yellow!25}440 & \cellcolor{green!25}23.5MB \\
    \textbf{Ours-small} & 0.779 & 0.286 & 27.00 & \cellcolor{red!25}361 & \cellcolor{red!25}25.8MB & 0.825 & 0.233 & \cellcolor{green!25}23.90 & \cellcolor{red!25}624 & \cellcolor{red!25}11.6MB & 0.894 & 0.273 & 29.24 & \cellcolor{red!25}568 & \cellcolor{red!25}12.3MB \\

    \bottomrule
    \end{tabular}
    }
\end{table*}

%% file: tables/pruning_level_comp.tex


\begin{table*}
    \caption{Proposed compression pipeline performance with various levels of pruning defined by $\gamma_{\text{iter}}$.}
    \label{tab:pruning_comp}
    \centering
    \scalebox{0.85}{
    \begin{tabular}{l|cccc|cccc|cccc}
    \toprule
           & \multicolumn{4}{c}{Mip-NeRF360} & \multicolumn{4}{c}{Tanks\&Temples} & \multicolumn{4}{c}{Deep Blending} \\
    \textbf{$\gamma_{\text{iter}}$} & \textbf{SSIM$^\uparrow$} & \textbf{PSNR$^\uparrow$} & \textbf{LPIPS$^\downarrow$} & \textbf{Mem}$^\downarrow$ & \textbf{SSIM$^\uparrow$} & \textbf{PSNR$^\uparrow$} & \textbf{LPIPS$^\downarrow$} & \textbf{Mem}$^\downarrow$ & \textbf{SSIM$^\uparrow$} & \textbf{PSNR$^\uparrow$} & \textbf{LPIPS$^\downarrow$} & \textbf{Mem}$^\downarrow$ \\
    \midrule
0.375 & 0.808 & 27.579 & 0.235 & 62.23MB & 0.845 & 24.059 & 0.191 & 27.90MB & 0.898 & 29.558 & 0.250 & 68.29MB \\
0.450 & 0.792 & 27.311 & 0.264 & 38.64MB & 0.838 & 24.084 & 0.209 & 18.82MB & 0.899 & 29.571 & 0.255 & 40.83MB \\
0.500 & 0.779 & 26.999 & 0.286 & 25.76MB & 0.825 & 23.903 & 0.233 & 11.65MB & 0.897 & 29.482 & 0.261 & 23.53MB \\
0.550 & 0.741 & 26.214 & 0.337 & 12.78MB & 
0.802 & 23.451 & 0.268 & 6.43MB & 0.894 & 29.241 & 0.273 & 12.32MB \\
0.600 & 0.692 & 25.142 & 0.394 & 5.73MB & 0.765 & 22.694 & 0.318 & 2.99MB & 0.887 & 28.646 & 0.294 & 5.40MB \\

    \bottomrule
    \end{tabular}
    }
\end{table*}

%% file: tables/modules_impact.tex
\begin{table}[]
\centering
\caption{The impact of each ELMGS module on the \texttt{garden} scene from the Mip-NeRF360 dataset, including Gradient and Opacity Aware Pruning (GAP), Learned Step size Quantization (LSQ), Morton Ordering (MO), and Entropy Coding (EC).}
\small 
\setlength{\tabcolsep}{3pt} 
\begin{tabular}{cccc|ccc} 
\toprule
GAP & LSQ & EC & MO & PSNR & Mem & Comp \\
\midrule
\ding{55} & \ding{55} & \ding{55} & \ding{55} & 27.296 & 1.4GB & 1$\times$ \\
\checkmark & \ding{55} & \ding{55} & \ding{55} & 27.026 & 306MB & 5$\times$ \\
\checkmark & \ding{55} & \checkmark & \ding{55} & 27.026 & 257MB & 5.5$\times$ \\
\checkmark & \ding{55} & \checkmark & \checkmark & 27.026 & 252MB & 5.5$\times$ \\
\checkmark & \checkmark	& \ding{55}	& \ding{55} & 27.023 & 306MB	& 5$\times$ \\
\checkmark & \checkmark & \checkmark & \ding{55} & 27.023 & 70MB & 19$\times$ \\
\checkmark & \checkmark & \checkmark & \checkmark & 27.023 & 64MB & 22$\times$ \\

\bottomrule
\end{tabular}
\label{table:post-hoc-pruning}
\vspace{-0.4cm}
\end{table}

%% file: display_figures/abalation_figures.tex
\begin{figure*}
    \begin{tabular}{@{}c@{}c@{}c@{}}
        {\includegraphics[width=0.3\linewidth]{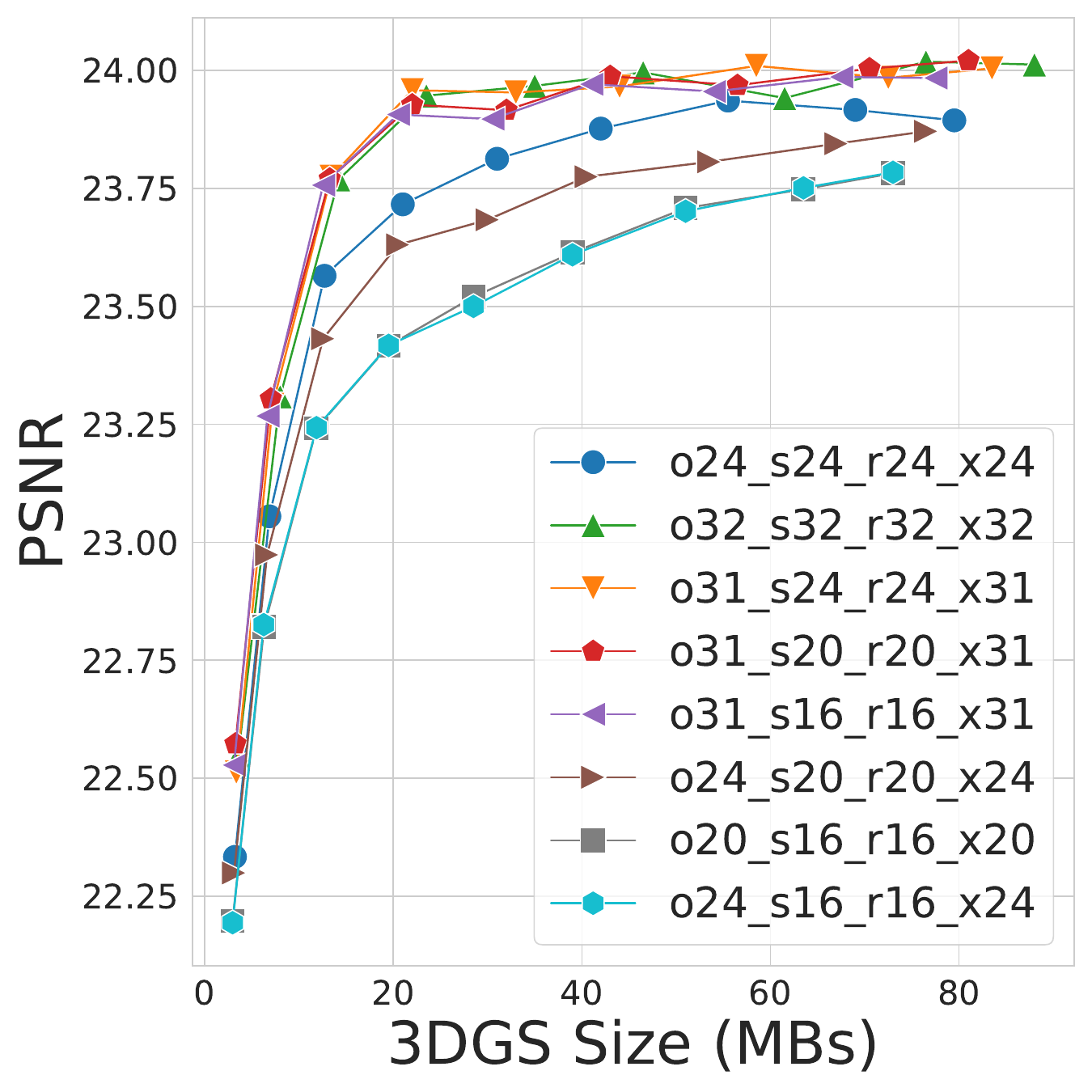}}&
        {\includegraphics[width=0.3\linewidth]
        {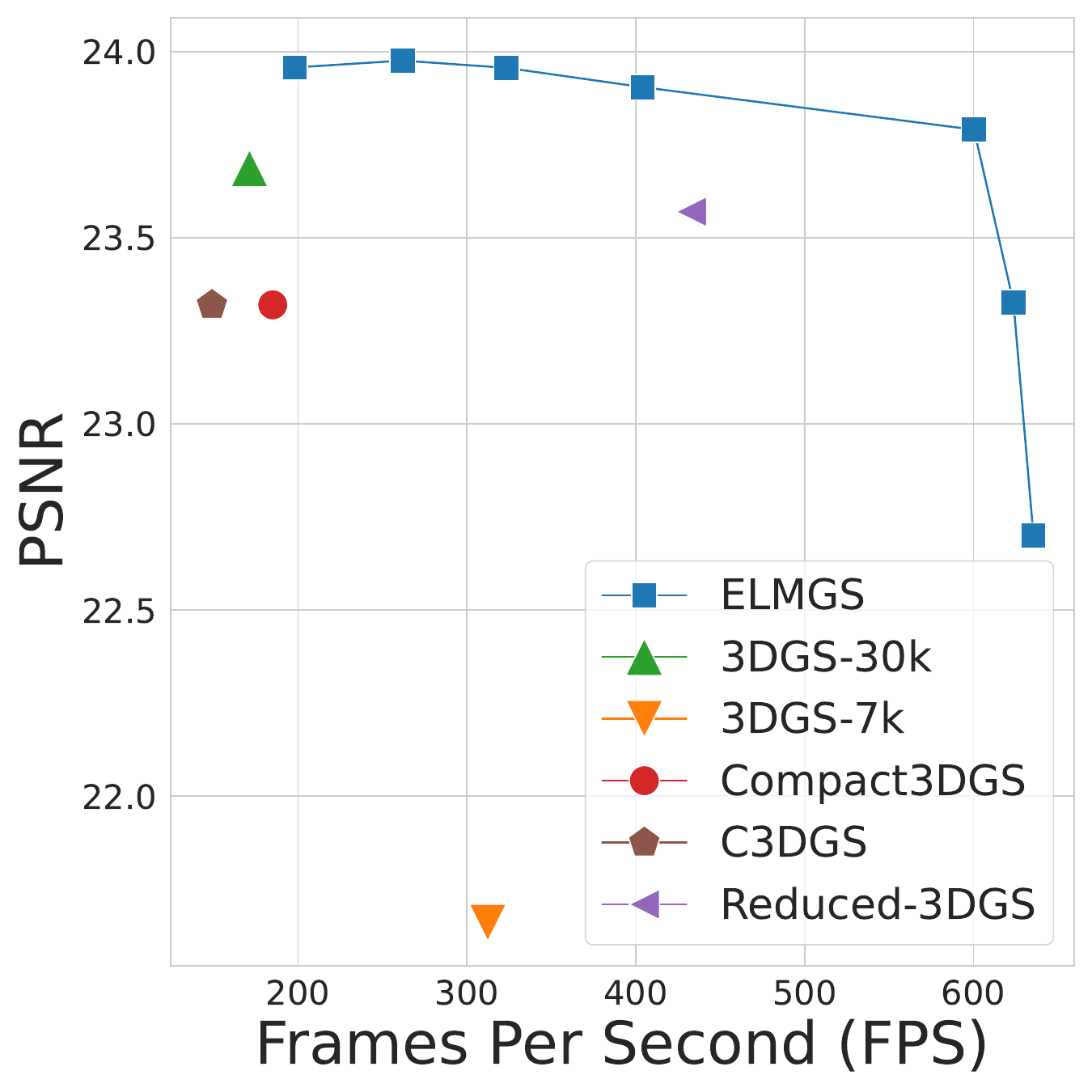}}&
        {\includegraphics[width=0.4\linewidth]{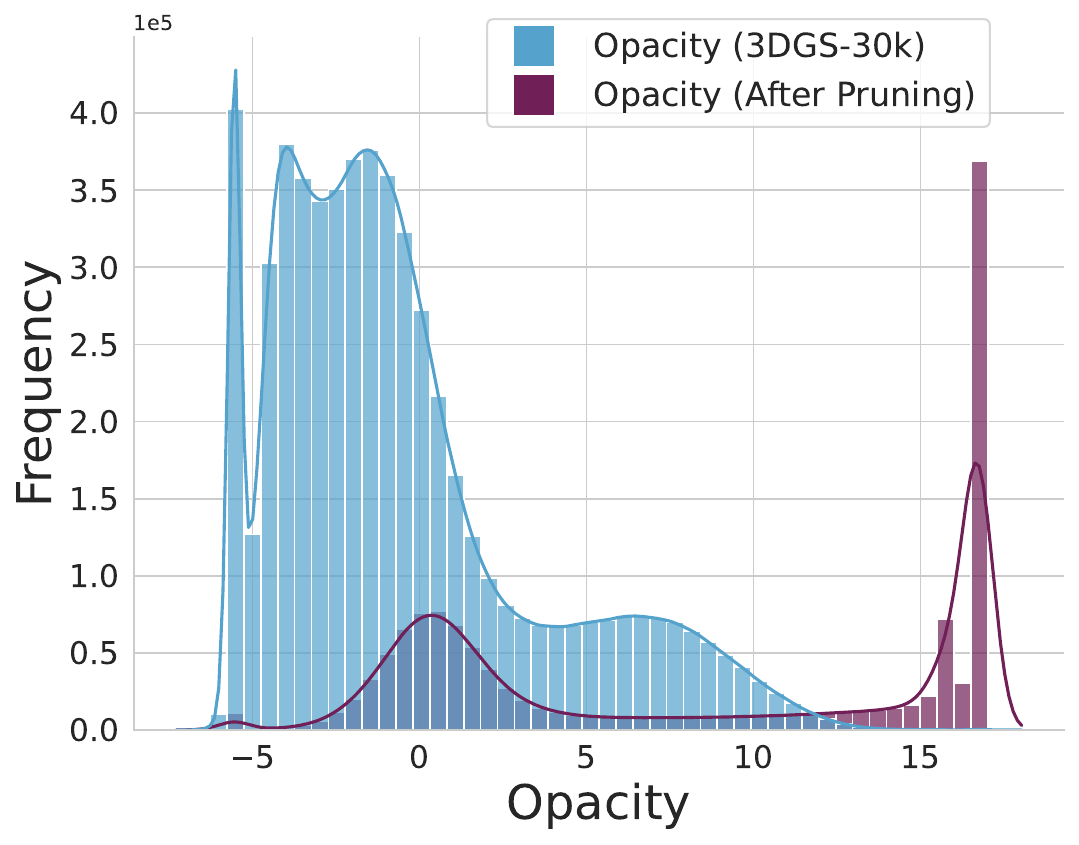}}\\
        (a)&(b)&(c)\\
    \end{tabular}
    \vspace{-10pt}
    \caption{{Tradeoff of performance and size at different bit-depths where opacity (o), scale (s), rotation (r), XYZ position (x) (a) and in terms of FPS on the Tanks\&Temples dataset (b), and opacity distribution before and after GAP for the \texttt{garden} scene  from the Mip-NeRF360 dataset (c).}}
    \label{fig:abl-studies}
\end{figure*}

%% file: 5_conclusion.tex
\vspace{-0.1cm}
\section{Discussion}
\label{sec:Discussion}
\salman{ELMGS is a flexible, plug-and-play method that can be applied to any 3DGS-based approach. With the widespread use of 3DGS in applications like inverse rendering~\cite{jiang2024gaussianshader,shi2023gir} and animatable avatars~\cite{lei2024gart}, various adaptations have been developed~\cite{fei20243d}. Some methods, such as Gaussian Shader~\cite{jiang2024gaussianshader} and GIR~\cite{shi2023gir}, introduce additional attributes or enhance the baseline algorithm, as demonstrated by Mini-Splatting~\cite{fang2024mini}. Other approaches, including Radsplat~\cite{niemeyer2024radsplat} and GART~\cite{lei2024gart}, make structural changes to the baseline 3DGS. ELMGS can be seamlessly integrated with any of these 3DGS variants to improve rendering speed and achieve higher compression ratios. Given the widespread use and versatility of 3DGS-based methods, our approach is effectively applicable across various implementations, owing to its plug-and-play nature and high performance.}

\section{Limitations}
\salman{ELMGS offers significant improvements in rendering speeds and achieves high compression gains while maintaining performance comparable to the baseline. However, one notable limitation is the extreme pruning, which allows us to significantly reduce the number of Gaussians only up to a certain threshold before performance starts to deteriorate. Additionally, our method requires training on top of a pre-trained 3DGS model, which increases the overall training cost. Future work will aim to enhance pruning ratios and reduce training times to address these limitations.}

\section{Conclusion}
We present ELMGS, a novel plug-and-play compression pipeline for 3DGS, incorporating Gradient and Opacity Aware Pruning (GAP), Learned Step size Quantization (LSQ), and entropy encoding with Morton ordering. Our method achieves compression rates of up to 38$\times$ and enhances rendering speed to over 600 FPS without compromising visual  quality on standard benchmark datasets. By substantially reducing the memory and computational footprint of 3DGS and adopting a hardware-friendly quantization approach, ELMGS paves the way for future research to utilize 3DGS on low-power AR/ VR and edge devices.

\label{sec:conclusion}